\documentclass{article}

\PassOptionsToPackage{numbers, compress}{natbib}
\usepackage[preprint]{neurips_2019}

\usepackage{amsmath,amssymb,amsthm} % math notation and theorem environments
\usepackage{xfrac}          % compact fractions
\usepackage[utf8]{inputenc} % allow utf-8 input
\usepackage[T1]{fontenc}    % use 8-bit T1 fonts
\usepackage{hyperref}       % hyperlinks
\usepackage{url}            % simple URL typesetting
\usepackage{booktabs}       % professional-quality tables
\usepackage{nicefrac}       % compact symbols for 1/2, etc.
\usepackage{microtype}      % microtypography
\usepackage{xcolor}         % comment text in different color
\usepackage{enumitem}       % customized enumerate environments
\usepackage[ruled,linesnumbered]{algorithm2e} % For formatting algorithms
\usepackage{subcaption}
\usepackage{multicol}

\newtheorem{theorem}{Theorem}

\newtheorem{corollary}{Corollary}
\theoremstyle{definition}
\newtheorem{definition}{Definition}

\title{DARC: Differentiable ARchitecture Compression}

\author{%
  Shashank Singh\thanks{This work was done while Shashank Singh was an intern at Amazon, New York.} \\
  Machine Learning Department\\
  Carnegie Mellon University\\
  \texttt{sss1@cs.cmu.edu} \\
   \And
   Ashish Khetan \\
   Amazon, New York \\
   \texttt{khetan@amazon.com} \\
   \AND
   Zohar Karnin \\
   Amazon, New York \\
   \texttt{zkarnin@amazon.com} \\
}

\newcommand{\argmin}{\operatornamewithlimits{argmin}}

\newcommand{\E}{\mathop{\mathbb{E}}} % Expectation operator
\renewcommand{\H}{\mathcal{H}} % Hypothesis class
\newcommand{\IID}{\stackrel{IID}{\sim}} % Independent and identically distributed
\renewcommand{\L}{\mathcal{L}} % Loss function
\newcommand{\R}{\mathbb{R}} % Real numbers
\newcommand{\X}{\mathcal{X}} % Covariate sample space
\newcommand{\Y}{\mathcal{Y}} % Response sample space
\newcommand{\Rad}{\mathfrak{R}} % Rademacher compexity
\renewcommand{\hat}{\widehat} % Make hats wide by default
\renewcommand{\tilde}{\widetilde} % Make tildes wide by default
\begin{document}

\maketitle

\begin{abstract}
    In many learning situations, resources at inference time are significantly more constrained than resources at training time.
    This paper studies a general paradigm, called Differentiable ARchitecture Compression (DARC), that combines model compression and architecture search to learn models that are resource-efficient at inference time. Given a resource-intensive base architecture, DARC utilizes the training data to learn which sub-components can be replaced by cheaper alternatives. 
    The high-level technique can be applied to any neural architecture, and we report experiments on state-of-the-art convolutional neural networks for image classification.
    For a WideResNet with $97.2\%$ accuracy on CIFAR-10, we improve single-sample inference speed by $2.28\times$ and memory footprint by $5.64\times$, with no accuracy loss.
    For a ResNet with $79.15\%$ Top1 accuracy on ImageNet, we improve batch inference speed by $1.29\times$ and memory footprint by $3.57\times$ with $1\%$ accuracy loss.
    We also give theoretical Rademacher complexity bounds in simplified cases, showing how DARC avoids overfitting despite over-parameterization.
\end{abstract}

\section{Introduction}
\label{sec:intro}

In machine learning, it is common that resources at inference time are significantly more constrained compared to training time.
For example, while neural networks used in computer vision and natural language processing are routinely trained using GPUs, trained networks are often deployed on embedded systems or mobile devices with limited memory and computational power.
As another example, it is common to train a model that will be applied continuously in a production setting; while training occurs for a limited time, the machine performing inference must run indefinitely. Thus, learning a more efficient model can reduce costs associated with hardware or energy usage.
As a result, numerous papers in the last few years have studied deep model compression and acceleration. Most of these papers provide either resource efficient model components \citep{jaderberg2014speeding,zhang2016accelerating,wu2017shift} or techniques for pruning parameters in unstructured or structured ways~\citep{lecun1990optimal,polyak2015channel,li2016pruning,he2017channel,luo2017entropy,luo2017thinet,zhuang2018discrimination}.

This paper proposes a general paradigm, called Differentiable ARchitecture Compression (DARC) for learning in the context of constrained resources at inference-time. Rather then suggesting a specific cheap component and using it blindly throughout a neural network, or trying to tune hyperparameters of layers as done in network pruning or quantization parameters, we take an approach inspired by Neural Architecture Search (NAS)~\citep{zoph2016neural,liu2018progressive,pham2018ENAS,kandasamy2018NAS,liu2018darts}. Our technique starts with a resource intensive network design, and based on the data, learns which components can be replaced with more efficient alternatives, while maintaining model output quality. The resource requirement of the final model is controlled via a regularization term; we show how this term should be defined depending on the objective, such as maximizing throughput or minimizing memory footprint.

DARC has a clear intuitive advantage when compared to methods that quantize parameters, or prune them in structured or unstructured ways. These approaches are inherently restricted in their search space. They cannot replace a layer with a structurally different layer, or replace two or more layers with a shallow alternative. For example, it might be the case that a convolutional layer cannot be pruned without hurting performance but can be replaced with a depth-wise separable convolution. Alternatively an LSTM layer might not be amenable to weight pruning, but could be replaced with a more efficient self-attention layer. The DARC algorithm, applied to deep networks, offers a way to obtain a rich search space in the context of model compression. The high-level idea is to partition the network into components, and explore alternatives to these components \emph{simultaneously}. 

Replacing all components simultaneously is crucial, as it provides a data-driven way to decide \emph{which} components can be replaced with cheap alternatives. Indeed, replacing layers blindly may sacrifice too much prediction performance, as can be seen in MobileNets~\citep{howard2017mobilenets,sandler2018mobilenetv2}, which exclusively use depthwise-separable convolutions; although computationally efficient, even the most accurate MobileNet model is far less accurate at ImageNet classification than, say, ResNet50.

We treat the core problem of learning good replacements for components as a sparse ensemble learning problem. This principled view allows us to draw guidelines from a simpler, analyzable cases, leading to an educated choice of regularization function and a simple, gradient-based learning scheme that avoids over-fitting. This learning scheme results in fast training, allowing us to apply our techniques directly on large-scale datasets such as ImageNet. This contrasts from most NAS methods, which first learn an architecture on a small dataset, and then fit this architecture on the larger dataset.

We present experiments on networks commonly used for computer vision applications. By applying our techniques on the ResNet Architecture~\citep{he2016resnets}, we breach the Pareto curve of state-of-the-art models on both ImageNet and CIFAR-10 datasets, in terms of accuracy vs.\ throughput and accuracy vs.\ model size. A few results from our compression framework: for a WideResNet model achieving $97.2\%$ accuracy on CIFAR-10, we improve single-sample inference speed by $2.28\times$ and memory footprint by $5.64\times$, with no loss in accuracy. For a modified ResNet50 model with $79.15\%$ Top1 accuracy on ImageNet, we improve inference speed by $1.29\times$ and memory footprint by $3.57\times$ with $1\%$ loss in accuracy. Both base models are publicly available from the GluonCV Model Zoo~\citep{ModelZoo}. Our experiment empirically demonstrates an intuitive observation that `you get what you optimize for', in that models minimizing model size tend to be quite different from those maximizing throughput.

We note that, while our experiments are limited to image classification architectures, the DARC method is applicable to any deep learning architecture, including models with recurrent cells or transformers, or indeed any sufficiently modular learning algorithm, as described in Section~\ref{sec:general_approach}, and any task with a well-defined objective function in which we would like to reduce inference costs. We see this work as a proof-of-concept for capabilities of the DARC framework, and the results in this paper give a strong indication that DARC can be applied on NLP architectures, or optimized in ways that we did not try here, e.g.\ for latency on devices other than GPU, or energy consumption.

\section{General Setting and DARC Estimator}
\label{sec:general_approach}

The intuition and motivation for our method starts with a task of model selection. Given a task and $J$ function families corresponding to candidate models, we are interested in finding the best model type for the task.
We relax this combinatorial optimization problem to a more tractable differentiable optimization problem (whence the name ``differentiable architecture compression''), by allowing convex combinations of these candidates.
This can be thought of as a constrained form of ensemble learning in which weights are restricted to represent a convex combinations of individual learners.

Then, we posit a budget constraint: each model type has an associated cost (e.g., memory consumption or latency), and the overall cost of the ensemble is the sum of costs of the used models.
Our task is then to learn a convex ensemble over a subset of candidates, with total cost within budget. We now formalize this approach, with modifications to address technical challenges as they arise.

In the sequel, for any positive integer $J$, $[J] = \{1,2,...,J\}$ denotes the set of positive integers at most $J$, and $\Delta^J := \left\{ x \in [0,1]^J : \sum_{j \in [J]} x_j = 1 \right\}$ denotes the probability simplex over $J$ elements.

Consider the conventional supervised learning setting, in which we have an i.i.d.\ training dataset $(X_1,Y_1),...,(X_n,Y_n) \IID P_{(X,Y)}$ from some joint distribution $P_{(X,Y)}$ on $\X \times \Y$. Fix a loss function $\L : \R \times \Y \to [0,\infty]$, and a hypothesis class $\H$ of $\R$-valued functions.
% \footnote{For simplicity we state our results with function of single outputs. We note that the framework applies for multiple value outputs such as multi-class classification tasks}.
We would like to learn a function $h : \X \to \R$, $h \in \H$ that minimizes the risk $R(h) := \E_{(X,Y) \sim P_{(X,Y)}} \left[ \L \left( h(X), Y \right) \right]$.
The usual empirical risk minimization (ERM) estimator is
$\hat h_{\text{ERM}} := \argmin_{h \in \H} \hat R(h)$ where, for any hypothesis $h \in \H$, $\hat R(h) := \frac{1}{n} \sum_{i = 1}^n \L \left( h(X_i), Y_i \right)$
denotes the empirical risk. To derive our resource-constrained objective, we impose a few structural assumptions on our hypothesis class $\H$:
\begin{enumerate}[label=\textbf{(A\arabic{*})}, ref=A\arabic{*},leftmargin=*,wide,itemsep=0pt]
    \item\label{assumption:convex_hull} $\H = \operatorname{Conv} \left( \bigcup_{j \in [J]} \H_j \right)$
    is the convex hull of a union of $J$ classes $\H_1,...,\H_J$.
    \item\label{assumption:costs} Each class $\H_j$ class has a known cost $C_j \geq 0$ of using a hypothesis $h_j \in \H_j$ at test-time.
    \item\label{assumption:additive_costs} Costs are additive: hypothesis $h = \sum_{j = 1}^J \alpha_j h_j \in \H$ has cost $C_{\ell_0}(\alpha) = \sum_{j = 1}^J C_j 1_{\{\alpha_j > 0\}}$.
    \item\label{assumption:budget} We have a known budget $B \geq 0$ for the final model at test-time.
\end{enumerate}

As we show in Section~\ref{sec:DARC_for_neural_networks}, these assumptions arise naturally in neural architecture compression. Given assumptions \eqref{assumption:convex_hull}-\eqref{assumption:budget}, the constrained ERM estimate is $\hat g = \sum_{j \in [J]} \hat \alpha_j \hat h_j$,
where
\begin{equation}
(\hat \alpha_0, \hat h_0) := \argmin_{\alpha \in \Delta^J,\hat h_j \in \H_j} \hat R \left( \sum_{j \in [J]} \alpha_j h_j \right),
  \quad \text{ subject to } \quad 
    C_{\ell_0}(\alpha) \leq B.
    \label{constraint:l0_cost}
\end{equation}
The above estimator is difficult (NP-hard) to compute, due to the non-smooth, non-convex budget constraint $C_{\ell_0}(\alpha) \leq B$. Since this constraint bounds the $\ell_0$ norm of $\alpha$ (weighted by $C$), the usual remedy would be to relax the constraint to one on the $\ell_1$ norm of $\alpha$ (weighted by $C$), namely $C_{\ell_1}(\alpha) := \sum_{j \in [J]} C_j \alpha_j \leq B$. Unfortunately, due to the constraint that $\alpha$ lies in the probability simplex $\Delta^J$ (which implies $\sum_{j \in [J]} \alpha_j = 1$), the $\ell_1$ constraint is insufficient to induce sparsity on $\alpha$.

Fortunately, sparse optimization on $\Delta^J$ is well-studied, with many solutions proposed, typically based on adding a smooth non-convex sparsity-inducing penalty~\citep{pilanci2012recovery,kyrillidis2013sparse,li2016methods}. Due to ease of implementation, we adopt a simple but effective solution proposed by \citet{kyrillidis2013sparse}, which involves alternating gradient updates with a projection operation $P_{\Delta^J} : \R^J\backslash\R_-^J \to \Delta_J$, given by
$P_{\Delta^J}(\alpha) = \frac{\alpha_+}{\|\alpha_+\|_1}$, where $\alpha_+ = (\max\{0,\alpha_1\},...,\max\{0,\alpha_j\}) \in \R_+$.

$P_{\Delta^J}$ is easy to compute, enforces the simplex constraint $\alpha \in \Delta^J$ exactly, and induces sparsity on $\alpha$.
For an intuition of how this works, one can note that $\nabla \|\alpha\|_2^2/2 = \alpha$, so that the update
$\alpha/\|\alpha\|_1 = \alpha - \left( 1 - 1/\|\alpha\|_1 \right) \alpha$
can be viewed as a gradient step for minimizing $-\|\alpha\|_2^2/2$ with adaptive step size $\left( 1 - 1/\|\alpha\|_1 \right)$.
As a technicality, we note that the projection $P_{\Delta^J}(\alpha)$ is undefined when $\alpha \in \R_-^J$ has no positive components. However, for realistic gradient step sizes $\eta$, this never occurs, since, after each gradient update, $\sum_{j \in [J]} \alpha_j \geq 1 - O(\eta)$.

Finally, a natural initial point for our procedure is one where $C_{\ell_1}(\alpha) > B$, hence we re-express the constraint $C_{\ell_1}(\alpha) \leq B$ as a penalty $\lambda C_{\ell_1}(\alpha)$. Since the value of $\lambda$ corresponding to $B$ is not known \emph{a priori}, we iteratively increase $\lambda$ until the solution of the optimization problem satisfies the budget constraint.
The resulting DARC procedure is shown Algorithm~\ref{alg:abstract_DARC}. We note that the ``stopping criterion'' for the inner loop can be as simple as a fixed number of training epochs (as in our experiments), or a more sophisticated early-stopping criterion.

\begin{algorithm}[H]
  \KwData{Training Data $\{(X_i,Y_i)\}_{i = 1}^n$, $J$ candidates $h_{1,w_1},...,h_{j,w_j}$ with initial parameters $w_1,...,w_j$ and costs $c_1,...,c_j \geq 0$, initial cost penalty parameter $\lambda_0 > 0$, budget $B$.}
  \KwResult{$\alpha,w_1,...,w_j$ such that $h = \sum_{j \in [J]} \alpha_j h_{j,w_j}$ has small risk $R(h)$ and cost $C_{\ell_0}(\alpha) \leq B$}
  $\alpha \leftarrow (1/J,...,1/J)$, $\lambda \leftarrow \lambda_0$ \;
  \While{$C_{\ell_0}(\alpha) > B$}{
    \While{stopping criterion is not met}{
      $(\alpha, w_1,...,w_J) \leftarrow (\alpha, w_1,...,w_J) - \eta \left( \nabla_{\alpha,w_1,...,w_J} \hat R \left( \sum_{j \in [J]} \alpha_j h_{j,w_j} \right) + C_{\ell_1}(\alpha) \right)$\;
      $\alpha \leftarrow P_{\Delta^J}(\alpha)$\;
    }
    $\lambda \leftarrow 2\lambda$
  }
 \caption{DARC algorithm for general hypotheses
    \label{alg:abstract_DARC}}
\end{algorithm}

% \begin{equation}
%     \hat h = \sum_{j \in [J]} \hat \alpha_j h_j,
% \quad\text{where}\quad
%     \hat \alpha := \argmin_{\alpha \in \Delta^J} \hat R \left( \sum_{j \in [J]} \alpha_j h_j \right),
% \quad\text{subject to}\quad
%     \sum_{j \in [J]} C_j \alpha_j \leq B,
%     \label{def:l1_DARC_estimator}
% \end{equation}
% where the $\ell_1$ is intended to induce sparsity on $\alpha$. This formulation is convex and differentiable almost everywhere, and is hence solvable by standard approaches. Unfortunately, however, the two constraints $\alpha \in \Delta^J$ (which implies $\|\alpha\|_1 = 1$) and $\sum_{j \in [J]} C_j \alpha_j \leq B$.

% The $\Psi:\R \to \R$ function above is a continuous function approximating the step function mapping $0$ to $0$ and $>0$ to $1$. For convex $\Psi$ the natural choice is the identity function, but from what we observe in what follows it is often beneficial to consider a non-convex alternative; the non-convexity may affect the training procedure but has the advantage of being closer to the desired cost.

\section{Applying DARC to Deep Networks}
\label{sec:DARC_for_neural_networks}

DARC can be applied in a myriad of ways to compress deep neural networks. In all of these ways, the basic premise is to intelligently replace various components of the network with cheaper components.

Consider a Neural Network (NN) with $L$ layers. For layer $\ell$, let $W_\ell$ be the parameters of the layer, and $g_\ell$ be the function mapping inputs and parameters to the output (in layers having no parameters, $W$ can be an empty token). For example, for a fully connected layer, $W$ is a matrix, the input $x$ is a vector, and $g$ is the matrix-vector multiplication function. We can write the NN as a function:
\begin{equation}
f(x) = g_L( W_L,  g_{L - 1}( W_{L - 1}, \cdots g_1( W_1, x) \cdots)),
    \label{eg:original_network}
\end{equation}
To apply DARC, we consider a set of replacement candidates $(g_{\ell,2}, W_{\ell,2}), \ldots, (g_{\ell,J_\ell}, W_{\ell,J_\ell})$ for each layer $\ell$ (with $g_{\ell,1}, W_{\ell,1}$ denoting the original function and weight of the layer). For each candidate $j$ in layer $\ell$, DARC takes as input an associated cost $C_{\ell,j} \geq 0$.
Examples of such costs include parameters count, FLOPs, or latency, which are usually easy to calculate or estimate experimentally.

Applying DARC to neural network compression then involves four main steps:
\begin{enumerate}[leftmargin=*]
    \item \textbf{Layerwise Continuous Relaxation:} First, we replace each $g_\ell, W_\ell$ with a weighted average
    \[\tilde{g}_\ell(\tilde{W}_\ell, \alpha_\ell, x) = \sum_{j=0}^{J_\ell} \alpha_{\ell,j} g_{\ell, j}(W_{\ell,j},x) \]
    where $\alpha_\ell \in \Delta^{J_\ell}$. The original network
    is replaced by $\tilde{f}(x) = \tilde{g}_L(\tilde{W}_L, \alpha_L, \cdots\tilde{g}_1(\tilde{W}_1, \alpha_1, x) \cdots )$.

    \item \textbf{DARC Model Initialization:} Before training the DARC model, we need to initialize the $\alpha$ weights and the parameters of the compression candidates. We initialized the $\alpha$ parameters as uniform vectors $\alpha_\ell = (1/J_\ell,1/J_\ell,...,1/J_\ell)$. The other option we considered was to put all weight on the original candidate ($\alpha_\ell = (1,0,...,0)$), so that the initial model was equivalent to the original model being compressed. However, this makes the gradient of the loss $0$ with respect to all parameters of the compression candidate, preventing these from training. Furthermore, the non-convex regularization discourages the weights of $\alpha_\ell$ to shift towards a value that makes use of the compression candidates. As for candidate parameters, we initialized each compression candidate to mimic the original layer, which we know gives good prediction results.
    In some cases, this can be done analytically (e.g.\ via PCA for lower dimensional fully-connected layers); more generally, this can be done via SGD, training the new candidate to minimize squared loss between its outputs and those of the original layer $g_{\ell,1}(W_{\ell,1},x)$. Since this is only for initialization, it suffices to use a small training sample and crude optimization procedure.
    
    \item \textbf{Training the Relaxed Model:} We minimize the empirical risk, simultaneously over the mixture weights ($\alpha$s) and the candidate weights ($\tilde W$s) as described in Algorithm~\ref{alg:abstract_DARC}.
    % , subject to the constraint $\tilde \Psi(C, \alpha) = \sum_\ell \sum_j C_{\ell,j} \Psi(\alpha_{\ell,j}) \leq B$, where $C$ and $\alpha$ are the vectors containing all the costs and weights of the components and $\tilde{\Psi}(C, \alpha)$ is a continuous approximation of the weighted $\ell_0$ loss $\sum_{\ell,j} C_{\ell,j} {bf 1}_{\alpha_{\ell,j}>0}$. Alternatively we may use a soft bound and add a regularization term of $\gamma \tilde{\Psi}(C, \alpha)$ to the optimization objective indicating the cost of the entire network. 
    
    \item \textbf{Selecting a Sub-Model:} Finally, in each layer $\ell$, we need to select a subset $S_\ell \subseteq [J_\ell]$ of compression candidates satisfying the original $\ell_0$ budget constraint. As discussed above, for sufficiently large $\lambda$, Algorithm~\ref{alg:abstract_DARC} converges to a solution with small (weighted) $\ell_0$ norm; i.e., $\alpha_\ell$ will have a small number of non-zero entries. Thus, we remove candidate $g_{\ell,j}$ (and its weight $\alpha_{\ell,j}$) from the network if $\alpha_{\ell,j} = 0$.
    % We note that, in realistic deep networks, obtaining meaningful compression or speedup usually requires $\|\alpha_\ell\|_0 = 1$ in each layer; in this case, the mixture weight $\alpha_\ell$ can be completely removed form the network.
    % , we gradually increase the regularization magnitude $\lambda$. For sufficiently large $\lambda$, the optimum will always lie near the boundary of $\Delta^J$ bounded weighted $\ell_0$ norm. This, combined with our projection step guarantees that the training procedure will converge to a subset $B$ of bounded $\ell_0$ cost.
    % \zk{Do you mean $\lambda$ term? Also, can we remove the comment about the proximal method?} \zk{Attempt for a replacement: As noted in Algorithm~\ref{alg:abstract_DARC}, we gradually increase the regularization magnitude $\lambda$. For sufficiently large $\lambda$, the optimum will always lie near the boundary of $\Delta^J$ bounded weighted $\ell_0$ norm. This, combined with our projection step guarantees that the training procedure will converge to a subset $B$ of bounded $\ell_0$ cost.}.

\end{enumerate}

An important note about the optimization procedure: Notice that we jointly optimize $\alpha$ and the model parameter on the same data, in contrast to other gradient based architecture search approaches that split the data into two training sets, optimize the model parameters on one and the $\alpha$ weights on the other. In Section~\ref{sec:theoretical_results} we analyze the Rademacher complexity of our procedure in a simplified setting and show that under the assumption that the original model defined by $g_{\ell,1}$ corresponds to a function family that is richer than the alternatives, optimizing all the parameters jointly does not hurt the generalization guarantees when compared to the original optimization objective where $J_\ell=1$. This statement holds naturally for model compression as opposed to an architecture search task.

\subsection{Efficient Approximate Convolutions}
\label{subsubsec:approximate_convolutions}

The computational demands of most deep networks used in computer vision problems, such as image classification, image segmentation, and object detection, are dominated by convolutional layers. Motivated by this, several papers have proposed efficient approximations to convolution, such as depthwise-separable convolutions~\citep{jaderberg2014speeding,zhang2016accelerating,howard2017mobilenets}, bottleneck convolutions~\citep{sandler2018mobilenetv2}, and shifts~\citep{wu2017shift}.

In a standard convolution layer we have $k \times k$ filters for every input and output channel. Denoting the output channels by $Y_i$ and the input channels by $X_j$, the $i$'th output channel is defined as $Y_i = \sum_j X_j * F_{i,j}$. Here $F_{i,j}$ is the appropriate filter and $*$ is the convolution operator. Restricting the discussion to the setting where the number of output and input channels are the same, a fully-grouped convolution is a more constrained alternative
% where $Y_i = X_i * F_i$
in which the filter is $k\times k$ but each output is computed based on a single input channel. A depthwise-separable convolution consists of a full-grouped convolution followed by a standard $1 \times 1$ convolution. In most setting this operation requires less compute and memory resources. A shift layer is an even cheaper alternative to depth-wise separable where the $F_i$'s are fixed and have only a single non-zero element, resulting in computational complexity equivalent to a single $1 \times 1$ convolution.
In what follows we use DARC to compress CNNs by considering alternatives from among the above options, for each convolution layer.

% \subsection{Pruning as a Special Case}
% \label{subsubsec:pruning}
% \SS{If time and space allow, mention that pruning can be expressed as a special case of DARC applied to each channel in a convolutional layer.}

\subsection{Other ways to compress CNNs with DARC}
\label{subsubsec:other CNN compres}
{\bf Channel Pruning}
Another obvious candidate for a cheap alternative to a convolution layer is a convolution layer that computes the output based on only a subset of inputs. We can consider all such possible alternatives and recover the objective of channel pruning. In order to avoid an exponential growth of $J$ we can restrict the alternatives to be based on a single input and obtain the standard framework for channel pruning where the choice of $\Psi$ results in some special regularization term aiming to minimize the number of input channels that we actually use.

{\bf Layer Pruning} could be done via an identity function candidate. Another option is to consider a single convolution, perhaps with enlarged kernel, to replace a sub-network of two or more convolutions.

\section{Theoretical Results}
\label{sec:theoretical_results}

In this section we provide results regarding the generalizability of a model learned via DARC. We restrict our attention to the simple case of learning an ensemble of models; as described below, the result has some implications for our algorithm for training DARC. This setting actually applies not only for DARC but for various Neural Architecture Search (NAS) methods such as DARTS~\citep{liu2018darts}, or ENAS~\citep{pham2018ENAS}. Indeed, at every level these papers aim to choose a single option or ensemble out of several options. While these methods differ in their approach to learning this ensemble, our generalization bound is independent of the learning technique.

Recall that DARC attempts to learn a convex combination of members of $J$ function families $\H_1,\ldots,\H_j$. Here, we analyze generalizability of this process via Rademacher complexity \cite{Bartlett:2003}, which is tightly connected with generalization bounds:

\begin{definition}[Rademacher Complexity]
Let $\H$ be a hypothesis class of functions mapping from $\X \to \R$ and let $n$ be an integer. Denote by $X_n$ a sample of i.i.d elements from $\X$ and $X_{n,i}$ the $i$'th sampled element. Denote by $\sigma$ an $n$ dimensional vector of i.i.d.\ uniform signs. The Rademacher complexity of $\H$ w.r.t. an integer $m$ is defined as
$\Rad_n(\H) =  \E_{\sigma, X_n}\left[ \sup_{h \in \H} \frac{1}{n} \sum_{i \in [n]} \sigma_i h(X_{n,i})  \right]$
\end{definition}

It is well known that the Rademacher complexity of a class $\H$ is equal to that of the convex hull of $\H$. Hence, in terms of Rademacher complexity, learning a convex combination is equivalent to learning a a single model from the $J$ different families of functions. We now move on to provide a generalization bound. To that end, recall that $R(h) = \E_{X,Y \sim D}[\L(h(X),Y)]$ denotes the distributional risk, and for a specific sample of $n$ data points $\{(X_i,Y_i\}_{i \in [n]}$, $\hat{R}(h) = \frac{1}{n} \sum_{i=1}^n \L(h(X_i),Y_i)$ denotes empirical risk over that training set. With $h=(h_1,\ldots,h_J) \in \H_1 \times \cdots \times \H_J$ and $\alpha \cdot h = \sum_j \alpha_j h_j$ we are ready for our generalization bound:
\begin{theorem}
Suppose we jointly estimate $\alpha,h_1,...,h_J$; i.e.,
\[\left( \hat \alpha, \hat h_1, ..., \hat h_J \right) := \argmin_{\substack{\alpha \in \Delta^J : C \cdot \alpha \leq B\\h_j \in \H_j}} \sum_{i = 1}^n \L \left( \alpha \cdot h(X_i), Y_i \right).\]
Let $\L(h(x),y) = 1_{\{f(x) \neq y\}}$ be $0$-$1$ loss. Then, with probability $\geq 1 - \delta$ (over $n$ training samples),
\[R \left( \hat \alpha \cdot \hat h \right) - \hat R \left( \hat \alpha \cdot \hat h \right) \leq \Rad_n \left( \bigcup_{j \in [J]} \H_j \right) + \sqrt{\frac{\log\sfrac{1}{\delta}}{n}}\]
\label{thm:joint_estimation}
\end{theorem}

Since Theorem~\ref{thm:joint_estimation} follows from straightforward application of standard Rademacher complexity-based generalization bounds (e.g., Theorem 5, part (b) of \citet{Bartlett:2003}), we omit its proof.

According to Theorem~\ref{thm:joint_estimation}, generalization error depends on a standard $\sqrt{(\log\sfrac{1}{\delta})/n}$ term and the Rademacher complexity of the union of function classes. If the function classes are diverse, this union can be quite rich and so $\Rad_n \left( \cup_{j \in [J]} \H_j \right)$ might be large, leading to overfitting. However, consider an example where $\H_1$ is the family of full convolutions, $\H_2$ is the family of depth-wise separable convolutions, $\H_3$ if the family of sparse convolutions, etc. Here, we actually have $\H_1 = \bigcup_{j \in [J]} \H_j$; thus, Rademacher complexity is simply that of the original model (i.e., with $J = 1$). Formally:
\begin{corollary}
Suppose that every sub-model is contained in $\H_1$; i.e., $\H_2,...,\H_J \subseteq \H_1$. Then, the Rademacher complexity of DARC is at most that of the original model: $\Rad \left( \H \right) \leq \Rad \left( \H_1 \right)$.
\label{corr:nested_models}
\end{corollary}
Even if we have $\H_1 \subsetneq \bigcup_{j \in [J]} \H_j $, in the setting of model compression the alternative families $\H_j, j>1$ are cheaper replacements for $\H_1$ so it stands to reason that $\Rad_n(\H_1) \approx \Rad_n(\bigcup_{j \in [J]} \H_j) $. This observation motivates our learning framework. It shows us that there is no need to split the training set, train the model parameters on one split and the control parameters on the other, as done in architecture search papers such as \citet{cai2018proxylessnas}, \citet{liu2018darts}.

We note that this result does not motivate a change in the learning framework of NAS. A key difference between NAS and DARC is that candidate models $\H_1,...,\H_J$ in NAS are intentionally diverse; their union is much richer than any individual. This translates to large $\Rad_n \left(\bigcup_{j \in [J]} \H_j \right)$, motivating a need to avoid jointly optimizing both $\alpha$ and model parameters on a single training set. When keeping a validation set aside for training $\alpha$, given the limited number of update steps typical in NAS papers, generalization error may be closer to the setting of fixed $\alpha$. In this setting, Rademacher complexity is bounded by $\sum_j \alpha_j \Rad_n(\H_j)$ \citep{cortes2014deep}, potentially much smaller than in Theorem~\ref{thm:joint_estimation}.

\section{Experimental Results}
\label{sec:experiments}

We applied DARC to a number of deep networks from the GluonCV Model Zoo~\citep{ModelZoo} for image classification on the CIFAR-10 and ImageNet~\citep{russakovsky2015imagenet} datasets; this section presents quantitative and qualitative results.
We report three performance metrics (model size, single-sample throughput, and batch throughput), and we specifically considered using DARC to minimize two of these (model size and batch throughput). Throughout this section, we use ``DARC(S)'' to denote DARC with a model (S)ize penalty, and we use ``DARC(T)'' to denote DARC with a batch (T)hroughput penalty.

\subsection{Implementation Details}

\paragraph{Choice of Compression Candidates}
For each convolutional layer, besides the original (full) convolution, we considered $3$ compression candidates: a (fully-grouped) depthwise-separable convolution with $3 \times 3$ kernels (abbreviated henceforth as ``3x3DS''), a full convolution with $1 \times 1$ kernels (``1x1FC''), and a $2$-layer candidate consisting of a 3x3DS layer followed immediately by a 1x1FC layer (``3x3+1x1'').
In ResNet50, which was already implemented using bottleneck convolutions~\citep{he2016resnets} consisting of a sequence of $1 \times 1$, $3 \times 3$, and $1 \times 1$ convolutions, the entire bottleneck sub-network was treated as a single component (i.e., all $3$ convolutions were replaced with a single block from the above mentioned alternatives).
We intentionally limited the choice of alternatives to maintain a simple system that can enjoy high throughput without special implementation. Our precise choice of alternatives is motivated by them already being an established component for deep networks, proven to work in some settings even without architecture search.

% We selected these candidates because they are easily implemented in most deep learning APIs. \SS{Is there a better justification we can give for these?}
% \SS{Actually, I briefly played around with a few other candidates (e.g., simple bottlenecks, identity layers, higher-rank grouped convolutions), but never saw these do well enough to justify keeping them. But I'm not sure whether these are actually bad candidates or if I just didn't investigate them enough. Should I mention these?} \zk{I think we can avoid mentioning these. }

\paragraph{DARC Training Details}
To maximize fairness when comparing with models in the GluonCV Model Zoo~\citep{ModelZoo}, most aspects of training DARC were based on the training scripts provided publicly by the Model Zoo\footnote{\href{https://gluon-cv.mxnet.io/_downloads/3bb06a6d6d085b1bb501b30aaf6c21c5/train_imagenet.py}{\texttt{train\_imagenet.py}}, \href{https://gluon-cv.mxnet.io/_downloads/b9ff6a58186006845fbf1a6e18d6c13e/train_mixup_cifar10.py}{\texttt{train\_mixup\_cifar10.py}}}. Due to space constraints, these implementation details are decribed in Appendix A; a few specific differences from these scripts are described below.

{\it Student-Teacher Initialization:}
As described in Section~\ref{sec:DARC_for_neural_networks} we initialized compression candidates to mimic original layers using student-teacher training. While this training had to be performed separately for each compression candidate in each layer of the original model, since each compression candidate has few parameters, each candidate's training converged quite quickly. Thus, in CIFAR-10 experiments, we simply ran $1$ epoch of the entire training dataset; in ImageNet experiments, we ran only $1000$ batches. A relatively large step size of $0.1$ was used, since teacher-student initialization is only for initialization and fine-tuning can be performed during model-selection.

{\it Main Training Phase:}
As described in Algorithm~\ref{alg:abstract_DARC},
training was conducted in blocks of epochs ($20$ epochs/block for CIFAR-10. $10$ epochs/block for ImageNet), with the compression penalty $\lambda$ increased after each block, to obtain a spectrum of compressed models.
For each dataset, penalty type, and model size, the initial value of $\lambda$ was selected to roughly balance the orders of magnitude of the empirical loss and the regularization term at the beginning of training.
After each block, we:
\vspace{-0.05in}
\begin{enumerate}[itemsep=0pt]
    \item Remove candidates $j$ with $\alpha_j = 0$ from the model.
    \item Save (for evaluation later) a copy of the DARC model, in which, in any layer with multiple non-zero $\alpha$ entries, all but the most expensive remaining candidate are removed.
    \item Decrease learning rate $\eta$ and increase compression penalty $\lambda$ (each by a factor of $2$).
\end{enumerate}
\vspace{-0.05in}
This blockwise training procedure was repeated until only one compression candidate per layer remained in the DARC model (this always happens eventually, as $\lambda$ increases). Finally, each saved model was fine-tuned for $20$ epochs using only prediction loss (i.e., without the compression penalty).
This procedure enabled us to obtain a sequence of compressed models at progressively increasing compression levels. Moreover, this ``warm-starting'' significantly improved compression speed since we only perform a total of $30$ epochs per $\lambda$ value, rather than the $>100$ epochs needed for convergence at high levels of compression.

\subsection{CIFAR-10 Results}

By all metrics, DARC gave the best results when applied to very wide models such as the WideResNet series (specifically, the WideResNet16\_10, WideResNet28\_10, and WideResNet40\_8 models~\citep{ModelZoo}). Moreover, unlike results on other ResNets, results on WideResNets were relatively similar for both DARC(S) and DARC(T); both versions of DARC selected the 3x3+1x1 candidate for every layer.
The reason for this is that very wide convolutions in WideResNet models can be replaced by depthwise separable convolutions with essentially no loss in accuracy, and significant improvement in throughput for both batches and for single samples ($1.4$-$1.6\times$ for each) and memory footprint ($4$-$6\times$). In the case of WideResNet16\_10, DARC produces a model with latency (single sample throughput) comparable to on of the fastest CIFAR-10 model (ResNet20\_v1; see Figure 1), while having accuracy within $0.8\%$ of the best model (ResNeXt29\_16x64d), which is $3.2\%$ accuracy points above the performance of ResNet20\_v1. For complete CIFAR-10 results see Appendix tables 3-4.

\subsection{ImageNet Results}

For ImageNet we compressed several ResNet models. To present the size compression results we provide Table~\ref{tab:resnet50} comparing accuracy change as a function of parameter reduction. We compared to previous published results compressing ResNet50 on ImageNet. To our knowledge these are the state-of-the-art results among those compressing ResNet50 on ImageNet. For compression as aggressive as X3, we maintain a drop of $0.86\%$ while the baseline suffers an accuracy drop of $3.26\%$.

For our throughput optimized models we could not find a published result on compressed ResNet models. We instead compare our pruned model to the linear interpolated Pareto Curve over the existing models; particularly relevant for us are the lines between ResNet18 and ResNet34, and between ResNet34 and Resnet50 models.
% \zk{Please mention how we got the Pareto curve that is used for the compressed ResNet34 models}
Table~\ref{tab:resnet34_50_pareto_throughput} provides throughput numbers for our compressed models alongside interpolated throughput of the Pareto Curve; our compressed models are well above the curve.

While further details, including other compressed versions, are available in the Appendix (Figures~2 and 3 and Table 5), Table~\ref{tab:resnet50} compares prediction performance of DARC(S) with that of state-of-the-art network pruning techniques applied to ResNet50 on ImageNet, at various compression levels.
In the light ($1.5 \times$) compression regime, the result of~\citep{zhuang2018discrimination} outperforms ours. The work of~\citep{zhuang2018discrimination} complements ours, in that their novelty is in warm-starting the compressed alternative not only to mimic the original, but also to be informative w.r.t.\ the label. Since this work does not have an architecture search component, we suggest that future work combine this clever warm-start with an architecture search component such as DARC. Once the compression becomes more aggressive, DARC outperforms the baseline, likely since in that regime (smaller network with more training epochs), a good architecture is more important than a good warm-start.
% \zk{If ok, I'd like to add the following: Notice that in the regime of light compression $1.5 \times$, the result of~\citep{zhuang2018discrimination} outperforms our own. We consider~\citep{zhuang2018discrimination} a complementary results to ours; the reason is that their novelty is in warm-starting the compressed alternative not only to mimic the original, but to be informative w.r.t.\ the label. Since this work does not have a component of architecture search, we consider a future work that combines the clever warm-start with an architecture search component such as DARC. Notice that once the compression becomes more aggresive, DARC outperforms the baseline, likely since in that regime having a correct architecture plays a more important role than a good warm-start. }

\setlength{\tabcolsep}{4pt}
\begin{table*}%[H]
\centering
\caption{}
\captionsetup[subtable]{position = below}
\begin{subtable}{0.52\linewidth}
   \centering
   \begin{tabular}{cp{2.3em}p{4.8em}p{4.5em}}
    Model            & Top1 & Throughput (1/256) & Pareto (1/256) \\
    \hline\hline
    ResNet34 (O) & 74.4 & 205 / 2442 & 205 / 2442 \\
    ResNet34 (T) & 73.9 & 234 / 2693 & 211 / 2569 \\
    ResNet34 (T) & 73.2 & 246 / 3289 & 220 / 2748 \\
    \hline
    ResNet50 (O) & 79.1 & 148 / 1242 & 148 / 1242 \\
    ResNet50 (T) & 78.3 & 176 / 1518 & 157 / 1447 \\
    ResNet50 (T) & 78.2 & 200 / 1603 & 159 / 1472 \\
    ResNet50 (T) & 76.8 & 208 / 1683 & 176 / 1829
    \end{tabular}
    \caption{Accuracy and throughput of models compressed by DARC(T) from ResNet34 and ResNet50 on ImageNet. (O)riginal denotes original model from GluonCV Model Zoo~\citep{ModelZoo}. (T)hroughput denotes models optimized by DARC to maximize throughput. ``Pareto'' is an estimate of throughput for the same accuracy, by linearly interpolating accuracy and throughput of ResNet34(O) and ResNet50(O).
    Throughput is in $224$px $\times 224$px images classified/second, in batch sizes $1$ and $256$.}
    \label{tab:resnet34_50_pareto_throughput}
\end{subtable}
\hspace*{1em}
\begin{subtable}{0.44\linewidth}
   \centering
		\begin{tabular}{lcc}
% 			\hline
    		    ResNet-50                             & Compression & Top1/Top5 \\
		    \hline\hline
    		    Disc~\citep{zhuang2018discrimination} & $1.51 \times$ & $\mathbf{+0.39}$/$\mathbf{+0.14}$ \\
    		    DARC(S)                               & $1.63 \times$ & $-0.51$/$-0.16$ \\
			\hline
    			ThiNet~\citep{luo2017thinet}          & $2.06 \times$ & $-1.87$/$-1.12$ \\
    		    Disc~\citep{zhuang2018discrimination} & $2.06 \times$ & $-1.06$/$-0.61$ \\
    		    DARC(S)                               & $2.05 \times$ & $\mathbf{-0.63}$/$\mathbf{-0.19}$ \\
			\hline
    		    Disc~\citep{zhuang2018discrimination} & $2.94 \times$ & $-3.26$/$-1.80$ \\
    	        DARC(S)                               & $3.01 \times$ & $\mathbf{-0.86}$/$\mathbf{-0.28}$ \\
		    \hline
    	        DARC(S)                               & $3.57 \times$ & $-0.99$/$-0.35$ \\
		    \hline
	            DARC(S)                               & $6.14 \times$ & $-6.09$/$-4.85$
		\end{tabular}
\caption{Comparison of DARC and state-of-the-art pruning methods when compressing ResNet-50 trained ImageNet for ResNet-50. (S)ize denotes models optimized by DARC to minimize model size. The Top1/Top5 accuracy of pre-trained model are $79.15\%$/$94.58\%$ respectively.}
\label{tab:resnet50}
\end{subtable}%
\vspace{-3ex}
\end{table*}

\subsection{Discussion of Compressed Architectures}
\vspace{-0.09in}

In both ResNets and WideResNets, model size tends to be dominated by a small number of the largest layers in the model, and is relatively insensitive to the depth of the network. Thus, significant compression can be achieved by replacing these large layers with 3x3+1x1 candidates, which offer compression of nearly $9 \times$ (for $3 \times 3$ convolutions). Since the sizes (number of convolutional kernels) of ResNet layers increases from bottom to top (i.e., from input to output) DARC(S) tends to first replace the top-most layers (i.e., layers closest to the output) of the network with 3x3+1x1 candidates, proceeding towards the bottom of the network as the compression parameter $\lambda$ increases.

In contrast, model latency is relatively uniformly distributed throughout the layers of the network -- the time taken to compute each convolutional layer scales only weakly with the number, size, and grouping of filters in that layer. Thus, in ResNets (but not in WideResNets) 3x3+1x1 candidates, which replace $1$ layer with $2$ smaller layers, tend to offer little or no benefit in throughput; of the compression candidates we considered, only the smallest (3x3DS) candidates offer significant acceleration (typically of about $2 \times$) over full convolutions. As a result, the acceleration offered by DARC(T) scales primarily with the number of layers that can be replaced by 3x3DS layers. Moreover, the replaced layers tend to be scattered throughout the network (rather than clustered near the output of the network). As noted earlier, in WideResNets, each layer is so large that the 3x3+1x1 candidate \textit{is} significantly faster than full convolution, and DARC(T) selects this candidate for every layer. 

Overall, we see that directly optimizing for size or for speed can lead to in significantly different compression strategies. This parallels recent work~\citep{cai2018proxylessnas} showing, in a NAS setting, that optimizing for speed on different hardware (GPU, CPU, or mobile) leads to different models.
While, in this paper, we focus on GPU throughput, we note that the computational speedup of a depth-wise separable convolution over a full convolution is typically even larger on CPU and embedded devices than on GPU devices. Thus, we expect that experiments similar to ours would provide efficient architectures for these alternative hardware environments.
% \zk{If ok, I'd like to add: In this paper, our focus was on GPU throughput. We would like to mention that for CPU and embedded device, the difference between a depth-wise separable convolution and a full convolution is typically larger than it is on GPU devices. As a result, we expect that experiments similar to ours to provide highly efficient architectures for these alternative hardware environments}

\section{Related Work}
\label{sec:related_work}

Work on accelerating and compressing deep neural networks has abounded in recent years, with the introduction of diverse techniques ranging from parameter pruning~\citep{lecun1990optimal,polyak2015channel,li2016pruning,he2017channel,luo2017entropy,luo2017thinet,zhuang2018discrimination}, low-rank factorization~\citep{jaderberg2014speeding,zhang2016accelerating,howard2017mobilenets}, fast approximate convolutions~\citep{bagherinezhad2017lcnn,wu2017shift},
knowledge distillation~\citep{hinton2015distilling,romero2014fitnets}, and quantization~\citep{gong2014compressing,han2015deep,zhou2017incremental,lin2016fixed}. See \citet{cheng2018model} for a survey of common approaches.

Compared to these methods, DARC has the advantage of a richer search space of alternative components, and the ability to replace multiple components by a single one (e.g.\ replacing a bottleneck sub-network of 3 convolutions with a single convolution). Moreover, DARC could be adapted to include these methods, though this is beyond the scope of our paper. Our suggestion of initializing compression candidates to mimic the original layer is also reminiscent of knowledge distillation.

Another closely-related line of papers concerns Neural Architecture Search (NAS) \citet{pham2018ENAS,kandasamy2018NAS,liu2018darts,gordon2018morphnet,cai2018proxylessnas}.
The most relevant papers are \citet{liu2018darts}, \citet{gordon2018morphnet}, and \citet{cai2018proxylessnas}, who all use a sparse linear component weighting scheme similar to DARC. \citet{liu2018darts} focused on pure architecture search, in which the goal is simply to find an architecture maximizing prediction performance, without consideration of inference-time efficiency. \citet{gordon2018morphnet} do not aim to replace layers with general alternatives but rather discover parameters in a data-driven way; their experiments are restricted to results in channel pruning. Recently, \citet{cai2018proxylessnas} performed NAS with a latency regularization term similar to ours.

Our methods differ from these NAS papers in two main ways. First, our architecture search is guided by an established base model that was already tested and proven useful. This distinction allows a simpler and more efficient learning scheme (motivated in Section~\ref{sec:theoretical_results}), which doesn't involve iterating back and forth between training sets, optimizing the model and $\alpha$ parameters. Furthermore, starting with a pre-trained model allows us not only to reach an effective architecture, but to warm start the weight parameters. As evidence of the advantage of starting with a pretrained model, our compressed model ResNet50(T) on ImageNet has Top1 accuracy $\geq 3\%$ more than models obtained by these NAS papers; thus, it seems that starting architecture search with a highly accurate base model can improve the efficiency/accuracy trade-off of the learned model.
% indeed, the objective in some of these papers is more strict in terms of throughput/memory, but it seems likely that in order to get an efficient accuracy model, our approach is superior to prior work.
Another distinction from gradient-based NAS results is our use of sparsity-inducing regularization; in previous papers the choice between the candidates is done via a softmax layer. This restricts the output to be a convex combination of inputs but does not optimize for sparsity. Given that they aim to achieve a sparse combination, they would likely benefit from a non-convex regularization term as in our paper.

Finally, a set of papers has studied generalization in ensembles of models. Our generalization bounds for DARC are similar to previous results for boosting and other ensemble prediction models~\citep{freund1997decision,cortes2014deep}. The most relevant result is a generalization bound of \citet{cortes2014deep} for a weighted mixture of $J$ classes given a fixed (data-independent) set of mixture weights; however, the assumption of a fixed set of mixture weights is too strong for our setting. Also, the bound of \citet{cortes2014deep} is for a $\rho$-margin loss and scales as $1/\rho$, whereas our result applies to $0$-$1$ loss (i.e., $\rho$-margin loss with $\rho = 0$).

\section{Conclusions and Future Work}
\label{sec:conclusion}

Empirically, we have shown that even a naive implementation of the DARC algorithm, utilizing only depthwise-separable approximations as compression candidates, can be used to compress large state-of-the-art deep neural networks, significantly improving both their inference speed and memory footprint. Moreover, intelligently making only some layers of the network depthwise-separable results in compressed models with significantly better predictive performance than simply making \textit{all} convolutions depthwise-separable, as in \citet{howard2017mobilenets}.

While depthwise-separable convolutions are easily implemented in existing deep learning packages and already enable substantial compression of many networks, future work may benefit from more sophisticated approximate convolutions, once they have efficient implementations. For example, shift operations~\citep{wu2017shift} are especially promising, as they require no stored parameters and replace expensive multiplication operations in convolution with fast indexing operations. Another venue worth pursuing is that of compressing a model into a shallower version. Although there are a few ways this could be attempted, such as an Identity candidate or replacing entire blocks of layers, it is unclear which technique would work best. Finally, we hope to apply DARC to architectures other than CNNs, and compress models with recurrent cells or transformers.

% \newpage
% \section{Citations, figures, tables, references}
% \label{others}

% \subsection{Tables}

% Note that publication-quality tables \emph{do not contain vertical rules.} We
% strongly suggest the use of the \verb+booktabs+ package, which allows for
% typesetting high-quality, professional tables:
% \begin{center}
%   \url{https://www.ctan.org/pkg/booktabs}
% \end{center}
% This package was used to typeset Table~\ref{sample-table}.

% \begin{table}
%   \caption{Sample table title}
%   \label{sample-table}
%   \centering
%   \begin{tabular}{lll}
%     \toprule
%     \multicolumn{2}{c}{Part}                   \\
%     \cmidrule(r){1-2}
%     Name     & Description     & Size ($\mu$m) \\
%     \midrule
%     Dendrite & Input terminal  & $\sim$100     \\
%     Axon     & Output terminal & $\sim$10      \\
%     Soma     & Cell body       & up to $10^6$  \\
%     \bottomrule
%   \end{tabular}
% \end{table}

\newpage
% \SS{Omitted for anonymity.}

{\small
\bibliographystyle{unsrtnat}
\bibliography{biblio}
}

\appendix
\section{DARC Implementation Details}
In this section, we provide further details about the implementation of DARC used in our experiments.

\paragraph{Environment Details}
We implemented DARC in Apache MXNet 1.3.1 using Python 3.6 and CUDA 9.0. Experiments were run on AWS EC2 p3.8xlarge and p3.16xlarge machines, which respectively features 4 and 8 NVIDIA Tesla-V100 GPUs. CIFAR-10 models were each trained with $1$ GPU. Smaller ImageNet models (ResNet18 and ResNet34) were trained using $4$ GPUs, while the larger ResNet50 was trained using $8$ GPUs.
% \zk{if we used 4 GPUs for ImageNet why did we use a p3.16xlarge and not p3.8xlarge? Or did we use 8 GPUs for ImageNet?} \SS{Actually, this isn't quite accurate. I used 8 GPUs (p3.16xlarge) for ResNet50, and 4 GPUs (p3.8xlarge) for the smaller models (ResNet18 and ResNet34)}.

\paragraph{Training Details}
Following the original script used to train models in the GluonCV Model Zoo, we utilized mixup training~\citep{zhang2017mixup}, and optimized cross-entropy loss with Nesterov accelerated stochastic gradient descent (NAG) with (default) momentum parameter $0.9$. As noted in the main paper, for student-teacher initialization, we used a relatively large learning rate $\eta = 0.1$. Thereafter, for model-selection, we began with an initial learning rate of $\eta = 0.01$, which was then halved after each traning block.

To minimize training time, training batch sizes were selected to be as large as possible without exceeding GPU memory during training. This resulted in batch sizes (per training GPU) of $256$ for ResNets on CIFAR-10, $128$ for WideResNets on CIFAR-10, $64$ for ResNet18 and ResNet34 on ImageNet, and $32$ for ResNet50 on ImageNet.

For each dataset, penalty type, and model size, the initial value of $\lambda$ was selected to roughly balance the orders of magnitude of the empirical loss and the regularization term at the beginning of training.
For CIFAR-10 experiments with size penalization, the initial value of the $\lambda$ compression penalty was set to $\lambda = 10^{-5} \times L$, where $L$ is the number of layers to which DARC was applied (i.e., the number of full convolutions in the original model). For CIFAR-10 experiments with latency penalization, we used $\lambda = 10^4 \times L$. For ImageNet experiments, we used $\lambda = 10^{-8} \times L$ with size penalization and $L = 10^4 \times L$ for latency penalization.
% \zk{Seems like a typo. Did you mean `and $\lambda = 10^4 \times L$ for latency penalization'?}.

\subsection{Measures of Model Performance}
\paragraph{Computational Performance}
As an estimate of model size, we report the size (on disk) of the parameter file created by MXNet when saving the model; this correlates well with both the number of parameters in the model and the footprint of the model in RAM or GPU memory.
Since throughputs are inherently noisy, we report average inference times over $1000$ batches.
Though multiple GPUs were used for training DARC, all inference times were computed using a single Tesla V100 GPU. We used batch size $1$ to estimate single-sample throughput and batch size $256$ to estimate batch throughput.
The cost of each compression candidate (i.e., number of parameters for DARC(S) or latency for DARC(T)) was calculated or estimated based on the student model trained during initialization.

\paragraph{Prediction Performance} On CIFAR-10, we used standard (``Top1'') prediction accuracy. On ImageNet, we additionally used (``Top5'') accuracy, the fraction of test images for which the correct label is among the five labels considered most probable by the model. We note that these are the standard performance used for these datasets~\citep{krizhevsky2010convolutional,krizhevsky2012imagenet}.
% \zk{Can we mention that (1) the loss being minimized by the optimization procedure is the standard cross-entropy loss (2) The performance measures we used are those commonly reported by papers that benchmark over CIFAR-10 and Imagenet.}

\section{Supplementary Results}

This section provides detailed numerical results of our experiments:

\begin{figure}[ht]
    \centering
    \includegraphics[width=\linewidth]{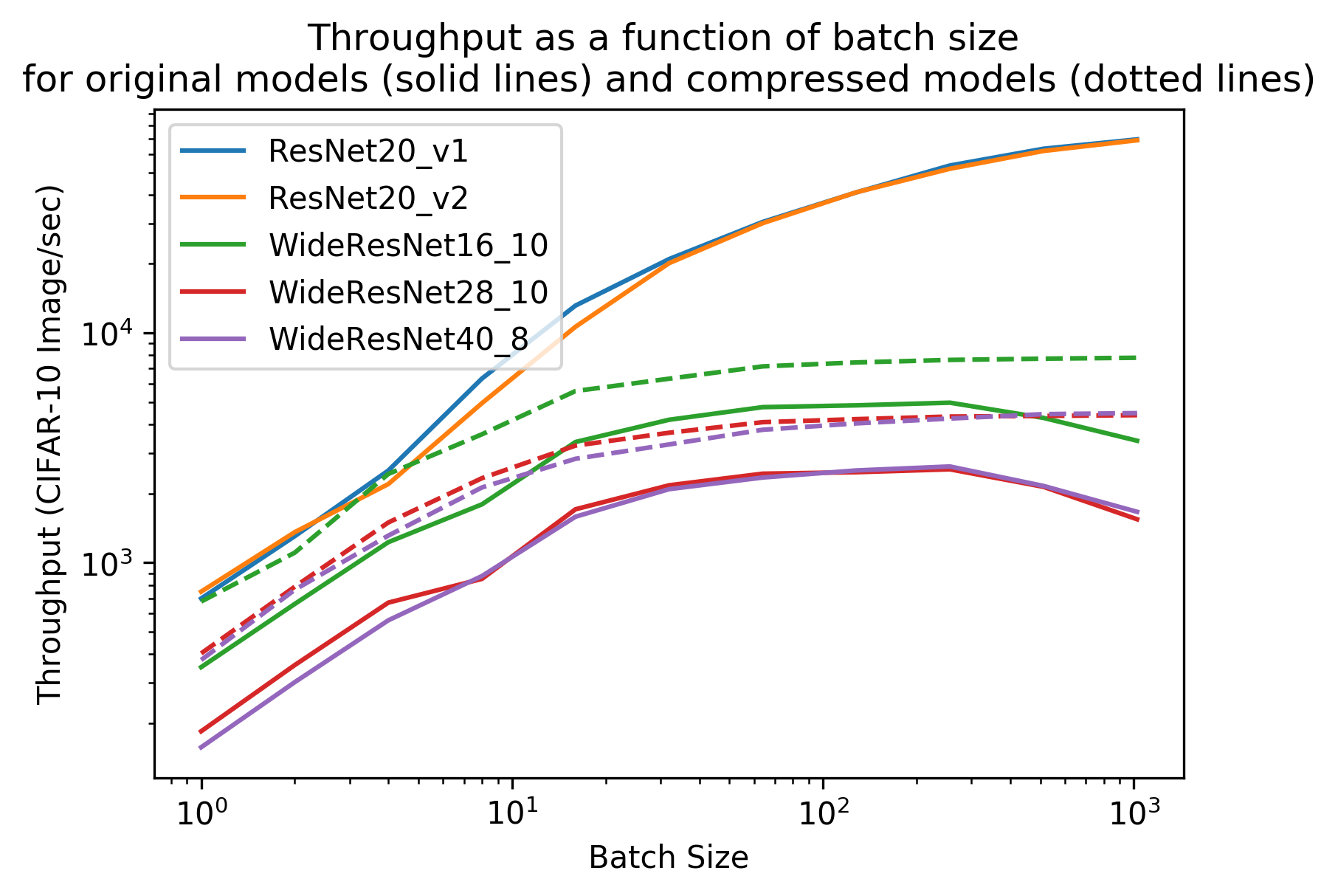}
    \caption{Throughputs of original and compressed WideResNet models, and of original ResNet20 models, at various batch sizes.}
    \label{fig:all_batch_sizes}
\end{figure}

\begin{table*}[ht]
    \centering
    \begin{tabular}{l|c|c|c|c}
        Model & Top-1 & Model Size & Throughput (1 Im/batch) & Throughput (256 Im/batch) \\
        \hline
        \hline
        ResNet20\_v1 (O) & 92.9 & 1.05 & 709.62 & 53020.76 \\
        ResNet20\_v1 (S) & 92.1 & 0.90 & 682.85 & 53078.90 \\
        ResNet20\_v1 (S) & 92.0 & 0.78 & 706.91 & 52822.23 \\
        ResNet20\_v1 (S) & 90.2 & 0.60 & 699.80 & 53206.18 \\
        ResNet20\_v1 (S) & 89.0 & 0.26 & 723.89 & 53058.42 \\
        \hline
        ResNet20\_v2 (O) & 92.7 & 1.05 & 792.03 & 52889.67 \\
        ResNet20\_v2 (S) & 92.1 & 0.85 & 665.98 & 53619.09 \\
        ResNet20\_v2 (S) & 91.4 & 0.67 & 671.05 & 54148.44 \\
        ResNet20\_v2 (S) & 91.0 & 0.48 & 700.55 & 54144.82 \\
        ResNet20\_v2 (S) & 90.3 & 0.26 & 663.76 & 56124.30 \\
        ResNet20\_v2 (S) & 89.1 & 0.25 & 739.73 & 55545.69 \\
        \hline
        ResNet56\_v1 (O) & 94.2 & 3.31 & 345.89 & 18930.70 \\
        ResNet56\_v1 (S) & 93.4 & 1.60 & 369.71 & 19974.62 \\
        ResNet56\_v1 (S) & 93.3 & 1.34 & 348.71 & 20330.08 \\
        ResNet56\_v1 (S) & 92.9 & 1.18 & 347.43 & 20711.30 \\
        ResNet56\_v1 (S) & 92.8 & 0.60 & 354.43 & 21155.00 \\
        ResNet56\_v1 (S) & 92.7 & 0.59 & 343.84 & 21648.25 \\
        \hline
        ResNet56\_v2 (O) & 94.6 & 3.30 & 395.35 & 19266.32 \\
        ResNet56\_v2 (S) & 93.4 & 1.12 & 382.69 & 20138.11 \\
        ResNet56\_v2 (S) & 93.2 & 1.12 & 360.47 & 20587.73 \\
        ResNet56\_v2 (S) & 93.1 & 0.93 & 353.93 & 20874.39 \\
        ResNet56\_v2 (S) & 92.9 & 0.71 & 355.93 & 20993.20 \\
        ResNet56\_v2 (S) & 92.7 & 0.63 & 335.27 & 21578.89 \\
        ResNet56\_v2 (S) & 92.5 & 0.59 & 330.72 & 22315.29 \\
        \hline
        ResNet110\_v1 (O) & 95.2 & 6.68 & 239.89 & 9704.98 \\
        ResNet110\_v1 (S) & 94.1 & 2.84 & 213.33 & 10391.16 \\
        ResNet110\_v1 (S) & 94.1 & 2.38 & 198.18 & 10704.93 \\
        ResNet110\_v1 (S) & 94.1 & 1.10 & 209.27 & 10858.91 \\
        ResNet110\_v1 (S) & 93.9 & 1.10 & 203.75 & 11084.18 \\
        ResNet110\_v1 (S) & 93.7 & 1.09 & 191.06 & 11245.70 \\
        \hline
        ResNet110\_v2 (O) & 95.5 & 6.68 & 220.21 & 9942.27 \\
        ResNet110\_v2 (S) & 94.3 & 3.13 & 210.58 & 10393.19 \\
        ResNet110\_v2 (S) & 94.3 & 2.08 & 209.11 & 10523.53 \\
        ResNet110\_v2 (S) & 94.2 & 1.97 & 199.03 & 10568.06 \\
        ResNet110\_v2 (S) & 93.9 & 1.21 & 214.44 & 10740.52 \\
        ResNet110\_v2 (S) & 93.7 & 1.14 & 219.05 & 11029.72 \\
        ResNet110\_v2 (S) & 93.7 & 1.09 & 209.54 & 11320.08 \\
        \hline
        WideResNet16\_10 (O) & 96.7 & 65.34 & 345.19 & 4914.29 \\
        WideResNet16\_10 (S) & 96.6 & 17.11 & 608.58 & 6932.60 \\
        WideResNet16\_10 (S) & 96.5 & 16.23 & 631.89 & 7546.51 \\
        \hline
        WideResNet28\_10 (O) & 97.1 & 139.24 & 186.63 & 2525.49 \\
        WideResNet28\_10 (S) & 97.2 & 26.42 & 423.48 & 3918.92 \\
        WideResNet28\_10 (S) & 97.1 & 24.67 & 425.66 & 4311.35 \\
        \hline
        WideResNet40\_8 (O) & 97.3 & 136.47 & 156.06 & 2617.89 \\
        WideResNet40\_8 (S) & 97.3 & 21.31 & 358.66 & 4258.24
    \end{tabular}
    \caption{Results of applying size-penalized DARC to CIFAR-10 models. (O)riginal Denotes an original model taken from the GluonCV Model Zoo~\citep{ModelZoo}. (S)ize denotes a model optimized by DARC to minimize model size. (T)hroughput denotes a model optimized by DARC to maximize model throughput. Model Size is provided in MB on disk. Throughput numbers are provided in $32$px $\times 32$px images/second.}
    \label{tab:cifar_size}
\end{table*}

\begin{table*}[ht]
    \centering
    \begin{tabular}{l|c|c|c|c}
        Model & Top-1 & Model Size & Throughput (1 Im/batch) & Throughput (256 Im/batch) \\
        \hline
        \hline
        ResNet20\_v1 (O) & 92.9 & 1.05 & 684.32 & 52239.12 \\
        ResNet20\_v1 (T) & 91.4 & 1.03 & 749.98 & 34743.19 \\
        ResNet20\_v1 (T) & 90.6 & 0.93 & 1144.10 & 48822.26 \\
        ResNet20\_v1 (T) & 88.3 & 0.92 & 1233.44 & 69868.65 \\
        \hline
        ResNet20\_v2 (O) & 92.7 & 1.05 & 737.07 & 52336.75 \\
        ResNet20\_v2 (T) & 91.9 & 1.01 & 691.04 & 60078.59 \\
        ResNet20\_v2 (T) & 90.7 & 0.83 & 1178.61 & 59269.61 \\
        ResNet20\_v2 (T) & 90.2 & 0.83 & 1275.66 & 70498.54 \\
        \hline
        ResNet56\_v1 (O) & 94.2 & 3.31 & 377.22 & 18966.31 \\
        ResNet56\_v1 (T) & 93.5 & 3.05 & 389.21 & 19175.07 \\
        ResNet56\_v1 (T) & 92.9 & 2.97 & 398.94 & 24985.82 \\
        ResNet56\_v1 (T) & 92.5 & 2.94 & 399.47 & 25682.29 \\
        \hline
        ResNet56\_v2 (O) & 94.6 & 3.31 & 389.76 & 18966.31 \\
        ResNet56\_v2 (T) & 94.4 & 3.16 & 398.09 & 20293.55 \\
        ResNet56\_v2 (T) & 94.3 & 3.16 & 409.56 & 20382.47 \\
        ResNet56\_v2 (T) & 94.3 & 3.15 & 409.78 & 21752.16 \\
        \hline
        ResNet110\_v1 (O) & 95.2 & 6.68 & 213.59 & 9615.24 \\
        ResNet110\_v1 (T) & 94.5 & 5.50 & 222.80 & 10161.07 \\
        ResNet110\_v1 (T) & 93.1 & 5.48 & 229.85 & 12565.72 \\
        ResNet110\_v1 (T) & 90.3 & 5.47 & 237.86 & 13032.99 \\
        \hline
        ResNet110\_v2 (O) & 95.5 & 6.68 & 174.33 & 9160.58 \\
        ResNet110\_v2 (T) & 94.9 & 5.35 & 225.93 & 10892.03 \\
        ResNet110\_v2 (T) & 94.5 & 5.20 & 226.46 & 11331.41 \\
        ResNet110\_v2 (T) & 94.3 & 5.19 & 233.13 & 11472.62 \\
        \hline
        WideResNet16\_10 (O) & 96.7 & 65.34 & 349.98 & 5019.10 \\
        WideResNet16\_10 (T) & 96.1 & 16.23 & 601.01 & 7699.13 \\
        \hline
        WideResNet28\_10 (O) & 97.2 & 139.24 & 158.52 & 2553.99 \\
        WideResNet28\_10 (T) & 96.8 & 24.67 & 455.45 & 4339.26 \\
        \hline
        WideResNet40\_8 (O) & 97.3 & 136.47 & 155.67 & 2643.25 \\
        WideResNet40\_8 (T) & 96.4 & 21.31 & 395.71 & 4269.20
    \end{tabular}
    \caption{Results of applying speed-penalized DARC to CIFAR-10 models. (O)riginal Denotes an original model taken from the GluonCV Model Zoo~\citep{ModelZoo}. (S)ize denotes a model optimized by DARC to minimize model size. (T)hroughput denotes a model optimized by DARC to maximize model throughput. Model Size is provided in MB on disk. Throughput numbers are provided in $32$px $\times 32$px images/second.}
    \label{tab:cifar_speed}
\end{table*}

\begin{table*}[ht]
    \centering
    \begin{tabular}{l|c|c|c|c}
    Model & Top-1 & Model Size & Throughput (1 Im/batch) & Throughput (256 Im/batch) \\
    \hline\hline
    ResNet18\_v1 (O) & 70.9 & 44.64 & 333.86 & 4044.52 \\
    ResNet18\_v1 (S) & 69.7 & 37.47 & 391.47 & 4142.90 \\
    ResNet18\_v1 (S) & 69.6 & 30.72 & 399.27 & 4196.04 \\
    ResNet18\_v1 (S) & 68.3 & 14.10 & 421.47 & 4281.12 \\
    ResNet18\_v1 (T) & 69.9 & 42.43 & 417.64 & 4296.34 \\
    ResNet18\_v1 (T) & 69.6 & 38.98 & 449.36 & 4473.29 \\
    ResNet18\_v1 (T) & 67.9 & 36.77 & 457.60 & 5173.91 \\
    \hline
    ResNet34\_v1 (O) & 74.4 & 83.23 & 205.15 & 2441.79 \\
    ResNet34\_v1 (S) & 73.5 & 37.68 & 216.68 & 2480.66 \\
    ResNet34\_v1 (S) & 72.4 & 31.56 & 220.29 & 2543.27 \\
    ResNet34\_v1 (T) & 73.9 & 70.66 & 233.92 & 2693.20 \\
    ResNet34\_v1 (T) & 73.2 & 60.77 & 245.57 & 3289.19 \\
    \hline
    ResNet50\_v1 (O) & 79.1 & 97.79 & 147.61 & 1242.42 \\
    ResNet50\_v1 (S) & 78.6 & 59.93 & 150.82 & 1257.26 \\
    ResNet50\_v1 (S) & 78.5 & 47.71 & 157.79 & 1263.03 \\
    ResNet50\_v1 (S) & 78.3 & 32.49 & 161.64 & 1298.97 \\
    ResNet50\_v1 (S) & 78.2 & 27.42 & 167.92 & 1359.69 \\
    ResNet50\_v1 (S) & 73.1 & 15.93 & 175.34 & 1492.43 \\
    ResNet50\_v1 (T) & 78.3 & 92.57 & 175.68 & 1518.25 \\
    ResNet50\_v1 (T) & 78.2 & 71.19 & 199.76 & 1602.78 \\
    ResNet50\_v1 (T) & 76.8 & 69.29 & 208.01 & 1682.92 \\
    \hline\hline
    ResNet101\_v1 (O) & 77.2 & 170.54 & 100.25 & 724.79 \\
    ResNet152\_v1 (O) & 78.1 & 230.49 & 69.27 & 500.19 \\
    \hline
    MobileNet1.0 (O) & 69.5 & 16.24 & 597.59 & 4393.48 \\
    MobileNet0.75 (O) & 66.2 & 9.94 & 627.24 & 6161.25 \\
    MobileNet0.5 (O) & 61.2 & 5.13 & 662.31 & 9805.38 \\
    \hline
    MobileNetV2\_1.0 (O) & 70.2 & 13.53 & 369.75 & 3615.16 \\
    MobileNetV2\_0.75 (O) & 68.1 & 10.15 & 376.56 & 4739.72 \\
    MobileNetV2\_0.5 (O) & 64.0 & 7.59 & 398.03 & 7065.31 \\
    \hline
    VGG11 (O) & 66.9 & 506.83 & 318.53 & 912.44 \\
    VGG13 (O) & 68.0 & 507.54 & 262.04 & 613.30 \\
    VGG16 (O) & 71.5 & 527.79 & 205.29 & 458.85 \\
    VGG19 (O) & 72.9 & 548.05 & 173.37 & 363.46 \\
    \hline
    DenseNet121 (O) & 74.0 & 30.80 & 119.88 & 1104.38 \\
    DenseNet161 (O) & 76.9 & 110.31 & 63.33 & 508.16 \\
    DenseNet169 (O) & 75.5 & 54.65 & 81.86 & 881.48 \\
    DenseNet201 (O) & 76.6 & 77.30 & 62.83 & 673.12
    \end{tabular}
    \caption{Results of applying DARC to ImageNet models. (O)riginal Denotes an original model taken from the GluonCV Model Zoo~\citep{ModelZoo}. (S)ize denotes a model optimized by DARC to minimize model size. (T)hroughput denotes a model optimized by DARC to maximize model throughput. Model Size is provided in MB on disk. Throughput numbers are provided in $224$px $\times 224$px images classified/second.}
    \label{tab:imagenet}
\end{table*}

\begin{figure}[ht]
    \centering
    \begin{subfigure}[b]{0.5\linewidth}
        \centering
        \includegraphics[width=\linewidth]{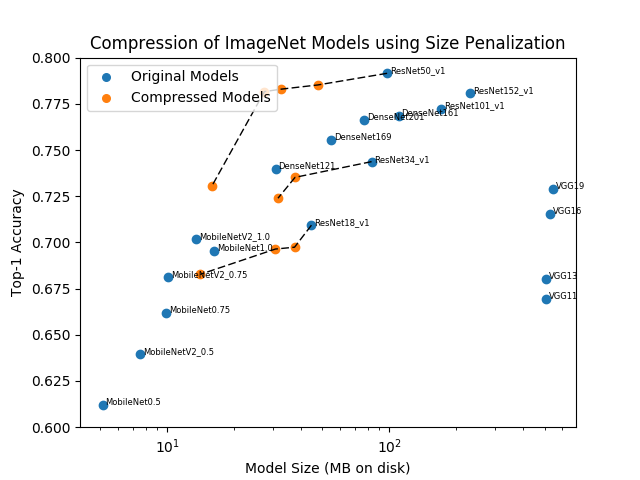}
    \end{subfigure}\\%
    \begin{subfigure}[b]{0.5\linewidth}
        \centering
        \includegraphics[width=\linewidth]{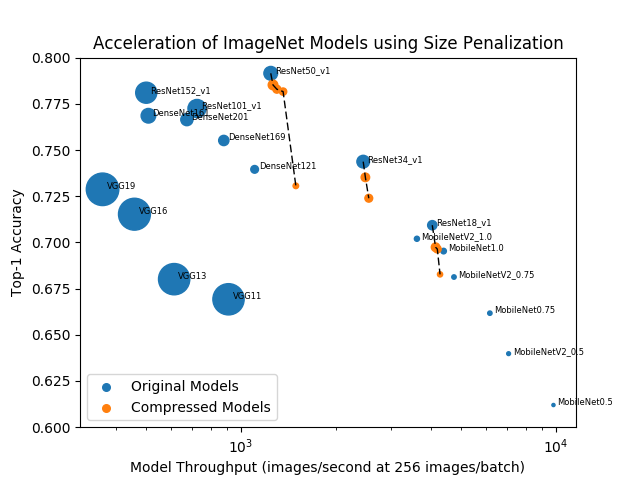}
    \end{subfigure}\\%
    \begin{subfigure}[b]{0.5\linewidth}
        \centering
        \includegraphics[width=\linewidth]{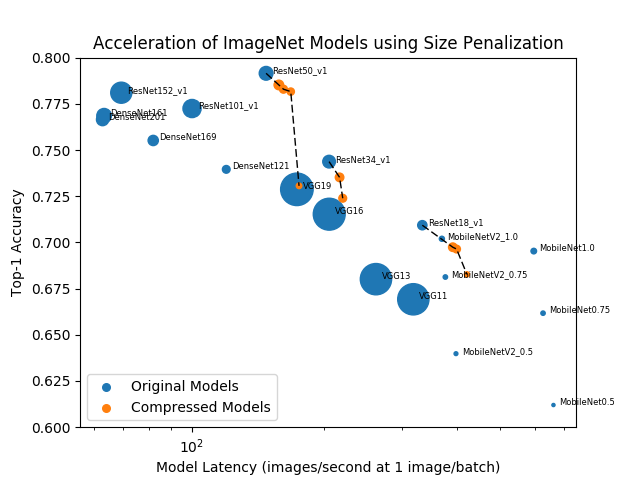}
    \end{subfigure}
    \caption{Compression of ImageNet models, in terms of size, throughput, and latency, when size is used as the computational penalty in DARC.}
    \label{fig:size_results_imagenet}
\end{figure}

\begin{figure}[ht]
    \centering
    \begin{subfigure}[b]{0.5\linewidth}
        \centering
        \includegraphics[width=\linewidth]{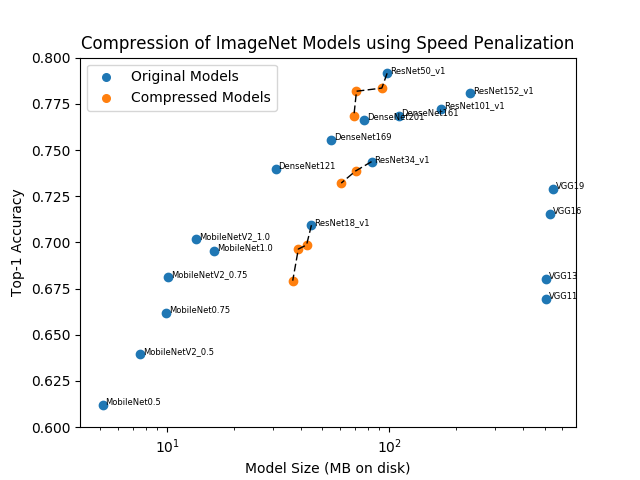}
    \end{subfigure}\\%
    \begin{subfigure}[b]{0.5\linewidth}
        \centering
        \includegraphics[width=\linewidth]{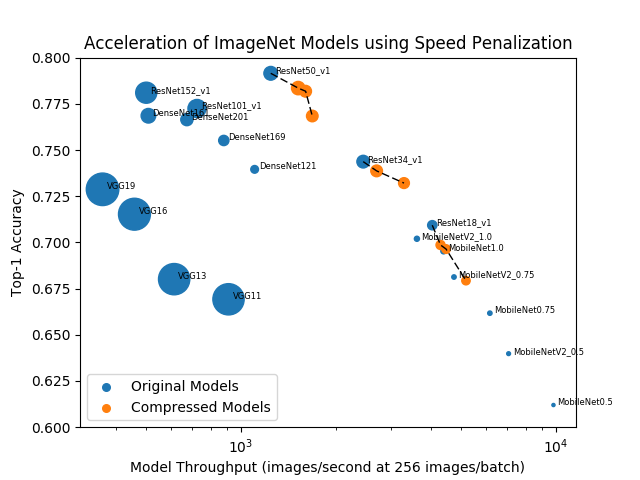}
    \end{subfigure}\\%
    \begin{subfigure}[b]{0.5\linewidth}
        \centering
        \includegraphics[width=\linewidth]{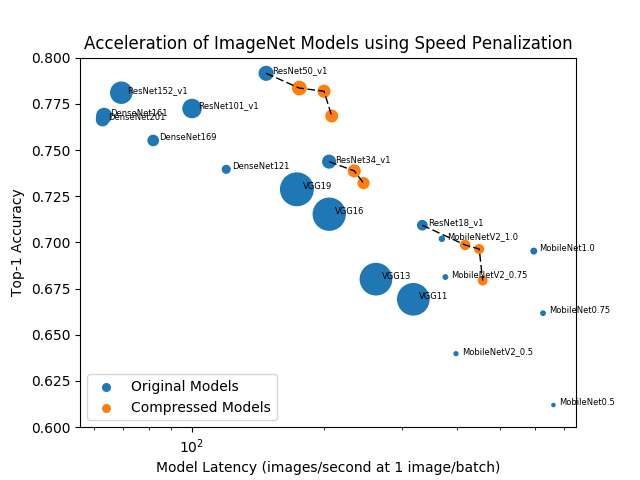}
    \end{subfigure}
    \caption{Compression of ImageNet models, in terms of size, throughput, and latency, when empirical latency is used as the computational penalty in DARC.}
    \label{fig:speed_results_imagenet}
\end{figure}

\end{document}